%% file: main.tex
\pdfoutput=1
\documentclass[runningheads]{llncs}

\usepackage[T1]{fontenc}
\usepackage{graphicx}
\usepackage{amsmath,amssymb}
\usepackage{array}
\usepackage{booktabs}
\usepackage{color}

\usepackage{hyperref}

\graphicspath{{figures/}}
\title{Did the Grid Erase the Event? EndoClock for Auditing Medical World-Model Pipelines\thanks{Accepted for publication at the 1st MICCAI Workshop on Medical World Models (MWM 2026).}}
\titlerunning{Did the Grid Erase the Event?}

\author{Yarin Udi\thanks{Corresponding author: \email{yarin@diastole.io}} \and Tom Sharon-Shahak \and Roee Masad \and Dan Pri-Tal}
\authorrunning{Y. Udi et al.}
\institute{Diastole Medical R\&D}  

\begin{document}
\maketitle

\input{sections/00_abstract}

\input{sections/01_introduction}
\input{sections/02_related_work}
\input{sections/03_setup}
\input{sections/05_diagnostic}
\input{sections/07_discussion}

\input{main.bbl}
\end{document}

%% file: sections/00_abstract.tex
\begin{abstract}
Medical world models commonly learn from multimodal recordings synchronized
onto a fixed-rate grid. This preprocessing resamples each native stream onto a
shared time axis. Each stream has an observation clock that governs when
observations are emitted or updated. When this clock depends on the latent or
acquisition state, it is \emph{endogenous}. In such settings, synchronization
may not be neutral and can erase task-relevant evidence before the model sees
the data.
We introduce a four-regime taxonomy that characterizes where the evidence
needed to distinguish a target event or state survives. The relevant witness
may remain in the sampled values, in grid-cell update patterns, in native
timing, or only in an external acquisition channel. \textsc{EndoClock}
operationalizes this taxonomy as a conservative pretraining audit. It reports
the lowest witness-bearing representation supported by the available evidence,
or unresolved when no regime can be established.
We illustrate this failure in echocardiography, where B-mode video write-outs
cease during pulsed-wave Doppler acquisition while the corresponding
measurement events remain recorded only in an external acquisition log.
This work is a preliminary failure alert and executable audit. Its practical
message is to preserve the native observation process long enough to determine
whether synchronization has erased information required by the intended task.

\keywords{Medical world models \and Multimodal fusion \and Identifiability
\and Endogenous sampling \and Observation-process audit.}
\end{abstract}

%% file: sections/01_introduction.tex
\section{Introduction}\label{sec:intro}

Medical world models commonly learn temporal dynamics from multimodal
recordings that have been synchronized onto a fixed-rate grid. This
preprocessing resamples each native stream onto a shared time axis. The
resulting representation also reflects which timing and acquisition channels
are retained alongside the resampled values. In applications such as
echocardiography, intensive-care monitoring, and wearable-plus-clinical fusion,
streams may pause, freeze, or change their sampling behavior with the
underlying physiological or acquisition state. In such settings,
synchronization may not be a neutral representation change and can remove
task-relevant evidence before the model sees the data.

Each native stream has an observation clock, which governs when
observations are emitted or updated. When this clock depends on the latent or
acquisition state, it is endogenous. Resampling may then preserve the
recorded values while erasing information carried by update patterns, event
timing, or acquisition-state transitions. The relevant question is therefore
not only which values were retained, but where the evidence needed to
distinguish the intended target event or state survives after preprocessing.

We refer to such evidence as a witness. It may survive in the sampled
values, in grid-cell update patterns, in native timing, or only in an external
acquisition channel. In the limiting case, every modeled stream is censored or
invariant while the distinguishing event is retained only outside the modeled
record. We call such an interval a \emph{no-witness interval}. Once this
representation has been formed, neither a finer grid nor any deterministic
augmentation of the modeled record can reconstruct the missing distinction.

We organize these possibilities into a four-regime taxonomy and operationalize
them through \textsc{EndoClock}, a conservative pretraining audit. It reports
the lowest witness-bearing representation supported by the available evidence,
or unresolved when the evidence is insufficient to establish a regime.
The audit should be applied early in data preparation, while native timestamps,
update patterns, and acquisition metadata are still available, so that
task-relevant evidence at risk of being discarded is identified before the
training representation is finalized.

\begin{figure}[ht]
\centering
\includegraphics[width=0.65\linewidth]{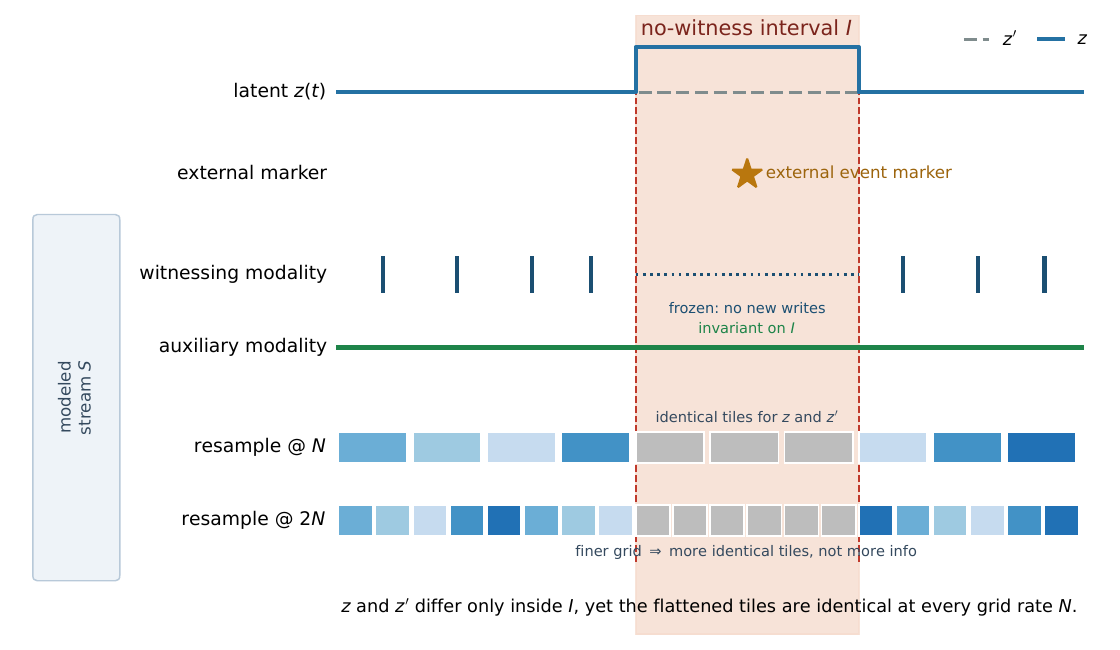}
\caption{\textbf{A no-witness interval.} Two latent histories differ only within $I$, yet every modeled
modality is censored or invariant there and the distinguishing marker lives only in an external side
channel. Fixed-grid preprocessing therefore produces the same modeled stream at any grid rate.}
\label{fig:nowitness}
\end{figure}

%% file: sections/02_related_work.tex
\section{Related work}\label{sec:related}

World models and temporal representation learners often operate on
fixed-rate sequential observations~\cite{ha2018world,hafner2019planet,hafner2025dreamerv3}. 
In medical applications, heterogeneous modalities such as images, physiological
waveforms, and device records are commonly aligned and fused before 
modeling~\cite{warner2024multimodal}. Medical video-synthesis models~\cite{reynaud2023echodiff}
and video-based JEPA representation learners~\cite{bardes2024vjepa} are likewise trained on
tensorized temporal inputs. 
Echocardiography-specific methods such as CoReEcho~\cite{maani2024coreecho} learn 
continuous 2D+time representations from synchronized clips. 
MRI CEKWorld~\cite{kong2026mricekworld} models continuous contrast-enhancement
dynamics from sparsely sampled MRI acquisitions using spatiotemporal consistency objectives.

These methods address how to learn from the representation supplied to the
model. Our question arises earlier, during representation construction, and
asks whether synchronization retained the task-relevant evidence in the first
place. Information discarded at this stage bounds what any downstream model
can recover, regardless of its architecture or capacity.

A complementary literature develops models for irregularly observed data.
Latent ODEs~\cite{rubanova2019latent}, neural CDEs~\cite{kidger2020neural}, 
GRU-D~\cite{che2018grud}, multi-time attention~\cite{shukla2021mtan}, 
and missing-data formulations~\cite{li2020learning} can exploit event timing, 
masks, and elapsed-time features that remain available in the modeled record. 
Classical missing-data theory formalizes the role of the observation 
mechanism~\cite{little2019statistical}, while work on informative observation in routine
healthcare data shows that measurement patterns may themselves carry
predictive information~\cite{sisk2021informative}.

When the task-relevant distinction survives in native event timing but not in
the fixed-grid values, these irregular-time models are the natural downstream
choice. 
EndoClock instead addresses the limiting case in which the witness required by
the target distinction is absent from the modeled record. In that case, masks,
timestamps, and $\Delta t$ features derived from the same record cannot reconstruct the
missing distinction (Sect.~\ref{sec:setup}), and the choice of
temporal model is immaterial. Conversely, equality of recorded values
alone is insufficient to establish failure, because event times, occupancy, or
reject patterns may still provide a witness.

Work on temporal coarsening shows that subsampling can alter
recoverable causal structure~\cite{gong2015discovery,hyttinen2016causal}.
EndoClock complements this literature with a task-specific pretraining audit 
for multimodal medical pipelines. 
Rather than proposing a new estimator, it checks whether the
representation supplied to a model still contains the evidence required to
distinguish the intended target event or state.

%% file: sections/03_setup.tex
\section{Problem setup and observation-clock taxonomy}
\label{sec:setup}

\paragraph{Observation process.}
Let $z:[0,T]\to\mathcal{Z}$ denote a latent physiological or acquisition-state
trajectory. Each modality $m$ produces observations
\[
S_m=\{(t_i^m,Y_i^m)\}_{i\geq 1},
\]
where $T_m=\{t_i^m\}$ is its native sequence of event times and $Y_i^m$ is the
corresponding observation. We call $T_m$ the modality's
\emph{observation clock}. The clock is \emph{endogenous} when its conditional
law depends on the latent or acquisition state. A practically important case
occurs when an acquisition-state transition pauses the modalities that would
otherwise reveal that transition.

Let $\mathcal{M}$ denote the modalities supplied to the model. External
acquisition channels are excluded from $\mathcal{M}$ unless they are explicitly
retained as model inputs. The native modeled record is
\[
S=(S_m)_{m\in\mathcal{M}}.
\]
A fixed-grid representation applies a deterministic resampling operator,
typically zero-order hold or interpolation,
\[
\hat S_N=R_N(S),
\]
on a grid of rate $N$.

For an interval $I=[a,b)$, let $S_I$ denote the restriction of $S$ to $I$, and
let $\mathcal{H}_a$ denote the observed history before $a$.

\begin{definition}[No-witness interval]\label{def:nw}
\normalfont
Let $z$ and $z'$ agree on $I^c$. The interval $I$ is a
\emph{no-witness interval} for the modeled record if
\[
\operatorname{Law}(S_I\mid z,\mathcal{H}_a)
=
\operatorname{Law}(S_I\mid z',\mathcal{H}_a)
\quad\text{a.s.\ in }\mathcal{H}_a.
\]
Thus, the modeled record has the same in-interval law under the two histories,
including both event times and recorded values.
\end{definition}

\paragraph{Observation-clock taxonomy.}\label{sec:taxonomy}
We now ask a broader question.
\emph{At what representation level does evidence for the target distinction
survive?} For a selected grid rate, we distinguish four regimes.

\begin{definition}[Observation-clock regimes]\label{def:taxonomy}
\normalfont
Fix a grid rate $N$ and a target distinction between latent histories $z$ and
$z'$ that agree on $I^c$. Write $V_N=R_N(S)$ for the fixed-grid values, let
$c(\cdot)$ range over cell-level functionals of the modeled record (occupancy,
update count, validity, reject pattern), and let
$T_I=(T_m\cap I)_{m\in\mathcal{M}}$ collect the native event times in $I$.
Each regime requires every coarser quantity to agree and the next one to differ.

\begin{itemize}
\setlength{\itemsep}{3pt}
\item[\textbf{(1)}] \textbf{Value-witnessed}, or clock-safe:
$\operatorname{Law}(V_N\mid z)\neq\operatorname{Law}(V_N\mid z')$.
The fixed-grid values already retain the distinction. This covers every case in
which the clocks are exogenous, or their state dependence is irrelevant to the
distinction.

\item[\textbf{(2)}] \textbf{Cell-witnessed}: $V_N$ has the same law under $z$
and $z'$, but $\operatorname{Law}(c(S)\mid z)\neq\operatorname{Law}(c(S)\mid z')$
for some cell-level functional $c$.

\item[\textbf{(3)}] \textbf{Sub-cell-witnessed}: $V_N$ and every $c(S)$ have the
same law, but the native event times do not:
\mbox{$\operatorname{Law}(T_I\mid z)\neq\operatorname{Law}(T_I\mid z')$}.
Only timing or ordering separates the histories.

\item[\textbf{(4)}] \textbf{External-witness-only}: $I$ is a no-witness interval
(Definition~\ref{def:nw}), so all of $S_I$ has the same law, but some external
channel $E\notin\mathcal{M}$ has
$\operatorname{Law}(E_I\mid z)\neq\operatorname{Law}(E_I\mid z')$ --- an
acquisition log, for example.
\end{itemize}
\end{definition}

The four regimes are mutually exclusive but not exhaustive. The audit returns
\textsc{unresolved} whenever the evidence establishes none of them
(Sect.~\ref{sec:diagnostic}). 

Table~\ref{tab:taxonomy} gives the lowest witness-bearing representation and 
recommended preprocessing for each regime.

\begin{table}[t]
  \centering
  \caption{\textbf{Observation-clock taxonomy and preprocessing policy.}
  The second column gives the lowest representation level that retains the
  target witness. The third gives the corresponding recommended preprocessing.
  When the audit establishes no regime it reports \textsc{unresolved}, whose
  policy is to preserve every native channel until the regime is resolved.}
  \label{tab:taxonomy}
  \footnotesize
  \setlength{\tabcolsep}{3pt}
  \begin{tabular}{@{}>{\raggedright}p{3.1cm}
                     >{\raggedright}p{3.2cm}
                     >{\raggedright\arraybackslash}p{5.2cm}@{}}
  \toprule
  \textbf{Regime} & \textbf{Witness-bearing level}
      & \textbf{Recommended policy} \\
  \midrule
  (1) Value-witnessed (clock-safe)
      & fixed-grid values
      & standard grid resampling is safe for this distinction \\[3pt]
  (2) Cell-witnessed
      & values $+$ update or validity summary
      & retain the update mask or validity flags alongside values \\[3pt]
  (3) Sub-cell-witnessed
      & values $+$ native timestamps or event ordering
      & retain native timestamps, or model the stream in continuous time \\[3pt]
  (4) External-witness-only
      & external acquisition channel
      & ingest the external channel, or redefine the target distinction \\
  \bottomrule
  \end{tabular}
\end{table}

\paragraph{Consequence of no-witness.}
Because every fixed-grid representation is a deterministic function of the
modeled record $S$, the regimes above are properties of that record rather than
of the grid. If two target histories induce the same law for $S$, then every
representation deterministically constructed from it --- any resampling operator
$R_N$ at any rate $N$, and any update mask, validity flag, time-since-update
feature $\Delta t$, or timestamp channel --- has the same law
under both, since each is a pushforward of the same input measure. Increasing
the grid rate or appending such features can therefore preserve a witness that
is already present (regimes~2--3), but cannot recreate one that is absent from
the modeled record (regime~4).

Full censoring is the clearest special case of a no-witness interval, but
modalities may also continue emitting while retaining identical event-time and
value laws. Equality of values alone is insufficient, because update timing may
itself remain a witness.

%% file: sections/05_diagnostic.tex
\section{\textsc{EndoClock}: audit and worked example}
\label{sec:diagnostic}

\textsc{EndoClock} operationalizes the taxonomy in
Table~\ref{tab:taxonomy} as a conservative pretraining audit. Given a target
distinction, candidate interval, modeled modalities, resampling operator, and
any available acquisition metadata, the audit works through increasingly
informative representations of the observation process. It first determines
whether the fixed-grid values retain the target distinction, or whether the clocks
can be certified as exogenous or task-irrelevant. It then examines cell-level
update and validity patterns, followed by native event timing and ordering.

\paragraph{Decision rules.}
\textsc{external\_witness\_only} is returned only when regime~4 is certified
\emph{structurally}. That requires zero write-outs on $I$ together
with a device-level rule that accounts for them, an invariant remaining
modality, and a verified external-only marker. A merely statistical failure to
reject equality of the competing observation kernels is not a proof of equality
and returns \textsc{unresolved}. The same holds for a non-detection of
clock-state dependence, which never certifies exogeneity. Symmetrically, a
modality that merely continues to update on $I$ is not a witness unless a
cell-level difference is actually exhibited, so it does not license a
\textsc{cell\_witnessed} verdict.

\paragraph{False positives and unresolved outcomes.}
Being conservative, the audit can flag an interval whose censoring is
state-correlated even though the target distinction does not depend on its
in-interval content. Such a false positive costs retained metadata, not lost
evidence. The converse error is excluded under full censoring, where the
in-interval record is empty under both histories and no modeled witness can
have been missed. By
design, \textsc{unresolved} is the default whenever certification fails,
directing the practitioner to preserve the native channels rather than to
conclude that resampling was safe.

\paragraph{Worked example: pulsed-wave Doppler acquisition.}
When an echocardiography console enters pulsed-wave Doppler (PWD) mode,
B-mode write-outs may cease while the operator holds the Doppler gate. In the
examined pipeline, PWD measurement events remain recorded in an external
acquisition log but are not retained in the modeled image stream.

\begin{figure}[ht]
\centering
\includegraphics[width=0.85\linewidth]{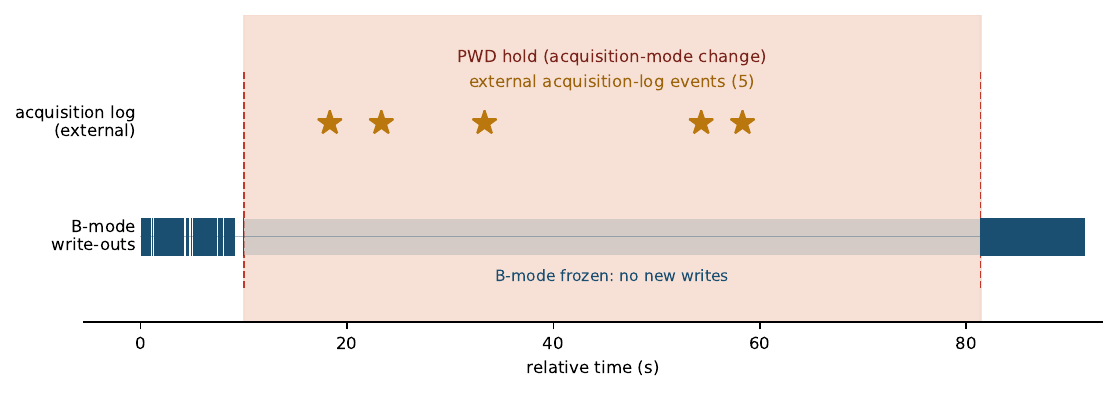}
\caption{\textbf{An external-witness-only interval in echocardiography.}
During the candidate interval, B-mode write-outs cease while five PWD
measurement events remain recorded in an external acquisition log. For the
modeled image stream, \textsc{EndoClock} returns
\textsc{external\_witness\_only}.}
\label{fig:pwd}
\end{figure}

The candidate interval was identified during routine preprocessing quality
assurance after an apparent mismatch between the continuously updating pose
clock and the frozen B-mode write-out stream. During the interval shown in
Fig.~\ref{fig:pwd}, the modeled image stream is fully censored, leaving no
cell-level or sub-cell image witness. Five PWD measurement events remain in the
log. \textsc{EndoClock} therefore returns
\textsc{external\_witness\_only} for this single de-identified acquisition.

\pagebreak
\paragraph{Example audit output.}
\vspace{-4pt}
\begin{center}\footnotesize
\begin{tabular}{@{}l@{\ \ }l@{\ \ }l@{}}
\texttt{any\_endogenous} & \texttt{True} & PWD-mode write-out suspension \\
\texttt{exogeneity\_certified} & \texttt{False} & endogenous, not certified safe \\
\texttt{all\_censored\_on\_interval} & \texttt{True} & zero B-mode write-outs on $I$ \\
\texttt{occupancy\_differs} & \texttt{None} & not testable, one observed history \\
\texttt{timing\_differs} & \texttt{None} & not testable, one observed history \\
\texttt{external\_marker\_only} & \texttt{True} & 5 PWD events, log only \\
\texttt{nonmarker\_clauses\_certified} & \texttt{True} & structural device rule, not a screen \\
\midrule
\multicolumn{3}{@{}l@{}}{$\Rightarrow$ \textsc{external\_witness\_only}} \\
\end{tabular}
\end{center}
\vspace{-6pt}

This example illustrates how the audit can identify a task-relevant witness
retained outside the modeled record. It does not establish the
prevalence of this failure mode or validate the audit across devices,
institutions, or acquisition workflows.

\paragraph{Controlled synthetic check.}
Beyond this single acquisition, we exercise all four regimes on a parameterized
generator with known ground truth. Within a constructed no-witness interval the
modeled values are label-independent by construction, so the evidential result
is not that a stream-only learner fails but that the failure is selective. 
From the same flattened stream, values-only recovers a coarse
target at $0.915$ AUC while the fine in-interval target stays at chance
($0.524$, $95\%$ CI $[0.482,0.564]$), and only the external marker fully
recovers it. The collapse persists across a $48\times$ grid-rate range
($5$--$240$\,Hz) and for every probe class we tried, from logistic regression to
a recurrent network.

%% file: sections/07_discussion.tex
\section{Conclusion}\label{sec:limits}

This paper examines a preprocessing failure mode in multimodal medical
world-model pipelines. Fixed-grid synchronization may erase task-relevant
evidence when observation clocks depend on physiological or acquisition state.
We introduced a four-regime taxonomy to characterize the lowest representation 
level at which the evidence required by a target distinction survives.

The relevance of observation-clock endogeneity depends on both the target
distinction and the representation supplied to the learner. State-dependent
sampling does not by itself imply failure, since the required witness may
remain in fixed-grid values, update patterns, or native timing. A representation
failure arises only when no modeled channel retains that distinction.

Such information loss occurs before model architecture or capacity becomes
relevant. Downstream temporal learners cannot recover evidence discarded
during representation construction. Practitioners should therefore define the
target distinction early and retain native timestamps, validity patterns, and
acquisition metadata until the representation has been audited. An external
witness should not automatically be added as a model input, because it may be
unavailable at inference, non-causal, or a source of target leakage.

Synchronization is therefore not always a neutral formatting operation.
Because deterministic processing cannot recreate a distinction absent from
the modeled record, increasing the grid rate does not resolve a true
no-witness interval.